%% file: main_arxiv.tex
\documentclass[conference]{IEEEtran}
\usepackage{cite}

\ifCLASSINFOpdf
\else
\fi
\usepackage{amsmath}
\usepackage{accents}
\DeclareMathOperator{\sgn}{sgn}
\newlength{\dhatheight}
\newcommand{\doublehat}[1]{%
	\settoheight{\dhatheight}{\ensuremath{\hat{#1}}}%
	\addtolength{\dhatheight}{-0.35ex}%
	\hat{\vphantom{\rule{1pt}{\dhatheight}}%
		\smash{\hat{#1}}}}
\usepackage{algorithm}
\usepackage[noend]{algpseudocode}
\def\BState{\State\hskip-\ALG@thistlm}

\ifCLASSOPTIONcompsoc
  \usepackage[caption=false,font=normalsize,labelfont=sf,textfont=sf]{subfig}
\else
  \usepackage[caption=false,font=footnotesize]{subfig}
\fi
\usepackage{url}

\usepackage{multirow}
\usepackage{booktabs}
\usepackage{tikz}
\usepackage{pgfplots}
\usepackage{tikzscale}
\usepackage{filecontents} 
\usepackage{rotating}

\usepackage{tabularx}

\begin{document}
%
\title{
	\begin{normalsize}
		\begin{flushleft}
			\copyright 2017 IEEE. Personal use of this material is permitted. Permission from IEEE must be obtained for all other uses, in any current or future media, including reprinting/republishing this material for advertising or promotional purposes, creating new collective works, for resale or redistribution to servers or lists, or reuse of any copyrighted component of this work in other works.\\
			\medskip
			This is a preprint version. The final version of this paper is available at \url{http://ieeexplore.ieee.org/document/8280935/} (DOI:10.1109/SSCI.2017.8280935).\\
		\end{flushleft}
		\hrulefill
		\color{white}
		.\\
	\end{normalsize}
	\color{black}
A Benchmark Environment Motivated by \\ Industrial Control Problems}



%
%
\author{\IEEEauthorblockN{Daniel Hein\IEEEauthorrefmark{1}\IEEEauthorrefmark{2}\IEEEauthorrefmark{4},
        Stefan Depeweg\IEEEauthorrefmark{1}\IEEEauthorrefmark{2}\IEEEauthorrefmark{4},
		Michel Tokic\IEEEauthorrefmark{2}, 
		Steffen Udluft\IEEEauthorrefmark{2},\\
		Alexander Hentschel\IEEEauthorrefmark{3}, 
		Thomas A. Runkler\IEEEauthorrefmark{1}\IEEEauthorrefmark{2} and
		Volkmar Sterzing\IEEEauthorrefmark{2}}
	\IEEEauthorblockA{\IEEEauthorrefmark{1}Technische Universit\"at M\"unchen, Department of Informatics, Boltzmannstr. 3, 85748 Garching, Germany}
	\IEEEauthorblockA{\IEEEauthorrefmark{2}Siemens AG, Corporate Technology, Otto-Hahn-Ring 6, 81739 Munich, Germany}
	\IEEEauthorblockA{\IEEEauthorrefmark{3}AxiomZen, 980 Howe St \#350, Vancouver, BC V6Z 1N9, Canada}
}


\maketitle

\input{abstract}


\newcommand\blfootnote[1]{%
  \begingroup
  \renewcommand\thefootnote{}\footnote{#1}%
  \addtocounter{footnote}{-1}%
  \endgroup
}
\blfootnote{\IEEEauthorrefmark{4} Equal contributions.}

%
\IEEEpeerreviewmaketitle

\input{introduction}

\input{rl-benchmarks}

\input{dynamics}

\input{setup}

\input{conclusion}

\section*{Acknowledgment}

The project this report is based on was supported with funds from the German Federal Ministry of Education and Research under project number 01IB15001. 
The sole responsibility for the report's contents lies with the authors.
The authors would like to thank Ludwig Winkler from TU Berlin for implementing the OpenAI Gym wrapper and sharing it with the community.



%
\bibliographystyle{IEEEtran}
\bibliography{bibliography}

\begin{appendices}

\input{states}

\input{goldstone}

\end{appendices}

\end{document}

%% file: abstract.tex
\begin{abstract}
In the research area of reinforcement learning (RL), frequently novel and promising methods are developed and introduced to the RL community.
However, although many researchers are keen to apply their methods on real-world problems, implementing such methods in real industry environments often is a frustrating and tedious process.
Generally, academic research groups have only limited access to real industrial data and applications.
For this reason, new methods are usually developed, evaluated and compared by using artificial software benchmarks.
On one hand, these benchmarks are designed to provide interpretable RL training scenarios and detailed insight into the learning process of the method on hand.
On the other hand, they usually do not share much similarity with industrial real-world applications.
For this reason we used our industry experience to design a benchmark which bridges the gap between freely available, documented, and motivated artificial benchmarks and properties of real industrial problems.
The resulting industrial benchmark (IB) has been made publicly available to the RL community by publishing its Java and Python code, including an OpenAI Gym wrapper, on Github.
In this paper we motivate and describe in detail the IB's dynamics and identify prototypic experimental settings that capture common situations in real-world industry control problems.
\end{abstract}

%% file: introduction.tex
\section{Introduction}

Applying reinforcement learning (RL) methods to industrial systems, such as in process industry like steel processing~\cite{schlang:99}, pulp and paper processing~\cite{runkler2003modelling}, and car manufacturing~\cite{hartmann2008online}, or power generation with gas or wind turbines~\cite{schaefer:07, schaefer:08}, is an exciting area of research. 
The hope is that an intelligent agent will provide greater energy efficiency and, desirably, less polluting emissions. However, the learning process also entails a significant amount of risk: 
we do not know beforehand how a particular learning algorithm will behave, and with complex and expensive systems like these, experiments can be costly. 
Therefore, there is high demand in having simulations that share some of the properties that can be observed in these industrial systems.

The existing simulation benchmarks have lead to great advancements in the field of RL. 
Traditionally simple dynamical systems like pendulum dynamics are studied, whereas nowadays  the focus has shifted towards more complex simulators, such as video game environments~\cite{bellemare13arcade}. 
Also in  the field of robotics very sophisticated simulation environments exist, on which new learning algorithms can be tested~\cite{todorov:12, lillicrap2015continuous}. 
The existence of such benchmarks  has played  a vital role in pushing the frontier in this domain of science.

For industrial control, however, such a test bed is lacking. 
In these systems we observe a combination of properties that usually are not present in existing benchmarks, 
such as high dimensionality combined with complex heteroscedastic stochastic behavior.
Furthermore, in industrial control different experimental settings are of relevance, for instance, the focus is usually less on exploration, and more on batch RL settings~\cite{lange2012batch}.

To this end, we recently developed the \textit{industrial benchmark}~(IB), an open source software benchmark\footnote{Java/Python source code: \url{http://github.com/siemens/industrialbenchmark}}, with both Java and Python implementations, including an OpenAI Gym wrapper, available. 
Previously, this benchmark has already been used to demonstrate the performance of a particle swarm based RL policy approach~\cite{hein2017batch}.
The contribution of this paper lies in presenting the complete benchmark framework as well as mathematical details, accompanied by illustrations and motivations for several design decisions.
The IB aims at being realistic in the sense that it includes a variety of aspects that we found to be vital in industrial applications like optimization and control of gas and wind turbines. 
It is not designed to be an approximation of any real system, but to pose the same hardness and complexity.
Nevertheless, the process of searching for an optimal action policy on the IB is supposed to resemble the task of finding optimal valve settings for gas turbines or optimal pitch angles and rotor speeds for wind turbines.

The state and action spaces of the IB are continuous and high-dimensional, with a large part of the state being latent to the observer. 
The dynamical behavior includes  heteroscedastic noise and a delayed reward signal that is composed of multiple objectives. 
The IB is designed such that the optimal policy will not approach a fixed operation point in the three steerings.
All of these design choices were driven by our experience with industrial challenges.

This paper has three key contributions: 
in Section~\ref{sec:comparison}, we will embed the IB in the landscape of existing benchmarks and show that it possesses a combination of properties other benchmarks do not provide, which makes it a useful addition as a test bed for RL. 
In Section~\ref{sec:description} we will give a detailed description of the dynamics of the benchmark. 
Our third contribution, described in Section~\ref{sec:Experimental_setup}, is to define prototype experimental setups that we find relevant for industrial control. 
Our goal is to encourage other researchers to study scenarios common in real-world situations.

%% file: rl-benchmarks.tex
\section{Placement of the Industrial Benchmark in the RL Benchmark Domain}\label{sec:comparison}

In the RL community numerous benchmark suits exist, on which novel algorithms can be evaluated. 
For research in industrial control we are interested in a particular set of properties, such as stochastic dynamics with high dimensional continuous state and action spaces. 
We argue that only few freely available benchmarks exist fulfilling these properties, thereby making our contribution, the
IB, a useful addition. To that end, we want to briefly review existing benchmarks in RL.

Classic control problems in RL literature~\cite{sutton:98}, such as the cart-pole balancing and mountain car problems, usually have low dimensional state and action spaces and deterministic dynamics. 
In the field of robotics more complex and high-dimensional environments exist with a focus on robot locomotion, such as  the MuJoCo environment~\cite{todorov:12,lillicrap2015continuous}. 
Other examples are helicopter flight\footnote{\url{https://sites.google.com/site/rlcompetition2014/domains/helicopter}}~\cite{abbeel_2010} or learning to ride a bicycle~\cite{Randlov+Alstrom:1998}. 
These systems, while complex, usually have deterministic dynamics or only limited observation noise.

Utilizing games as RL benchmarks recently brought promising results of deep RL into the focus of a broad audience. 
Famous examples include learning to play Atari games\footnote{\url{https://github.com/mgbellemare/Arcade-Learning-Environment}}~\cite{bellemare13arcade,Mnih2015} based on raw pixels, achieving above-human performance playing Ms.~Pac-Man~\cite{seijjen:17}, and beating human experts in the game of Go~\cite{silver_mastering_2016}. 
In these examples, however, the action space is discrete and insights from learning to play a game may not translate to learning to control an industrial system like gas or wind turbines.

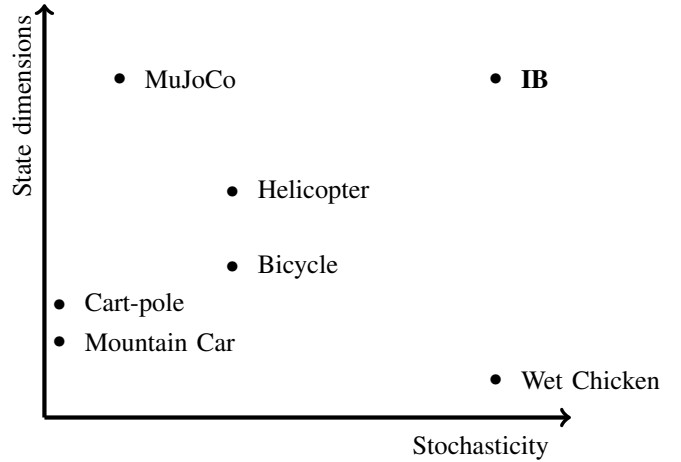
\begin{figure}[t]
	\begin{tikzpicture}
	\draw[->,ultra thick] (0,0)--(0,5.5) node[left,rotate=90,yshift=0.3cm]{State dimensions};
	\draw[->,ultra thick] (0,0)--(7,0) node[below,yshift=-0.1cm,xshift=-1.2cm]{Stochasticity};
	
	\node[label=right:{\textbf{IB}}]at(6,4.5){$\bullet$};
	\node[label=right:{Helicopter}]at(2.5,3){$\bullet$};
	\node[label=right:{MuJoCo}]at(1,4.5){$\bullet$};

	\node[label=right:{Cart-pole}]at(0.2,1.5){$\bullet$};
	\node[label=right:{Bicycle}]at(2.5,2.0){$\bullet$};
	\node[label=right:{Mountain Car}]at(0.2,1){$\bullet$};
	\node[label=right:{Wet Chicken}]at(6,0.5){$\bullet$};
	\end{tikzpicture}			
	\caption{
		Qualitative comparison of different RL benchmarks with continuous actions.
	 	The state space of the wet chicken 2D benchmark~\cite{hans:09} is rather low, but it is highly stochastic which makes it a challenging RL problem.
		Cart-pole and mountain car are deterministic benchmarks with few state dimensions and only a single action variable.
		The bicycle benchmark introduces some noise to simulate imperfect balance.
		The helicopter software simulation has a 12-dimensional state space and a 4-dimensional continuous action space.
		Stochasticity is introduced to simulate wind effects on the helicopter.
		The state space of the IB is high, since multiple past observations have to be taken into account to approximate the true underlying Markov state.
		Stochasticity is not only introduced by adding noise on different observations, but also by stochastic state transitions on hidden variables.
	}
	\label{fig:comp_sd_st}
\end{figure}

In Figure~\ref{fig:comp_sd_st} we give a qualitative overview on the placement of the proposed IB with respect to other RL benchmarks for continuous control. 
Here, we focus on stochasticity and dimensionality of the benchmarks at hand. Note that by stochasticity we do not only refer to the signal to noise ratio, but also to the structural complexity of the noise, such as heteroscedasticity or multimodality.
We conclude that the IB is a useful addition to the set of existing RL benchmarks. 
In particular, Figure~\ref{fig:comp_sd_st} illustrates that the combination of high dimensionality and complex stochasticity appears to be novel compared to existing environments.
In the following section, a detailed description and motivation for the applied IB dynamics is presented.

%% file: dynamics.tex
\section{Detailed description}\label{sec:description}
At any time step $t$ the RL agent can influence the environment, i.e., the IB, via actions $a_t$ that are three dimensional vectors in $[-1,1]^3$.
Each action can be interpreted as three proposed changes to the three observable state variables called current steerings.
Those current steerings are named velocity $v$, gain $g$, and shift $h$. Each of those is limited to $[0,100]$ as follows:
\begin{eqnarray}
a_t & = & (\Delta v_t,\Delta g_t, \Delta h_t),\\
\nonumber\\
v_{t+1} & = & {\rm max}(0,{\rm min}(100,v_t + d^\text{v}\Delta v_t)),\\
g_{t+1} & = & {\rm max}(0,{\rm min}(100,g_t + d^\text{g}\Delta g_t)),\\
h_{t+1} & = & {\rm max}(0,{\rm min}(100,h_t + d^\text{h}\Delta s_t)),
\end{eqnarray}
with scaling factors $d^\text{v}=1$, $d^\text{g}=10$, and $d^\text{h}=5.75$. The step size for changing shift is calculated as $d^\text{h} = 20 \sin(15^0)/0.9\approx 5.75$.

After applying action $a_t$, the environment transitions to the next time step $t+1$ 
in which it enters internal state $s_{t+1}$.
State $s_t$ and successor state $s_{t+1}$ are the Markovian states of the environment that are only partially observable to the agent.

An observable variable of the IB, setpoint $p$, influences the dynamical behavior of the environment but can never be changed by actions. 
An analogy to such a setpoint is, for example, the demanded load in a power plant or the wind speed actuating a wind turbine. 
As we will see in the upcoming description, different values of setpoint $p$ will induce significant changes to the dynamics and stochasticity of the benchmark. 
The IB has two modes of operation: 
a) fixing the setpoint to a value $p=const$, thereby
acting as a hyperparameter or 
b) as a time-varying external driver, making the dynamics become highly non-stationary.
We give a detailed description of setting b) in Subsection~\ref{setpoint_dyn}.

The set of observable state variables is completed by two reward relevant variables, consumption $c_t$ and fatigue $f_t$.
In the general RL setting a reward $r_{t+1}$ for each transition $t\rightarrow t+1$ from state $s_t$ via action $a_t$ to the successor state $s_{t+1}$ is drawn from a probability distribution depending on $s_t$, $a_t$, and $s_{t+1}$.
In the IB, the reward is given by a deterministic function of the successor state $r_{t+1}=r(s_{t+1})$, i.e.,
\begin{equation}
\label{eq:reward}
r_{t+1} = -c_{t+1} - 3f_{t+1}.
\end{equation}
In the real-world tasks that motivated the IB, the reward function has always been known explicitly.
In some cases it itself was subject to optimization and had to be adjusted to properly express the optimization goal.
For the IB we therefore assume that the reward function is known and all variables influencing it are observable.

Thus the observation vector $o_t$ at time $t$ comprises current values of the set of observable state variables, which is a subset of all the variables of Markovian state $s_t$, i.e.,
\begin{enumerate}
\item the current steerings, velocity $v_t$, gain $g_t$, and shift $h_t$,
\item the external driver, setpoint $p_t$,
\item and the reward relevant variables consumption $c_t$ and fatigue $f_t$.
\end{enumerate}
Appendix~\ref{appendix:states} gives a complete overview on the IB's state space.

The data base for learning comprises tuples $(o_t,a_t,o_{t+1},r_t)$. 
The agent is allowed to use all previous observation vectors and actions to estimate the Markovian state $s_t$.

The dynamics can be decomposed into three different sub-dynamics named operational cost, mis-calibration, and fatigue.

\subsection{Dynamics of operational cost}

The sub-dynamics of operational cost are influenced by the external driver setpoint $p$ and two of the three steerings, 
velocity $v$ and gain $g$. 
The current operational cost $\theta_t$ is calculated as
\begin{equation}
\theta_t = \exp\left(\frac{2 p_t + 4 v_t + 2.5 g_t}{100}\right).
\end{equation}
The observation of $\theta_t$ is delayed and blurred by the following convolution:
\begin{eqnarray}
\theta_t^\text{c} = \frac{1}{9} \theta_{t-5} + \frac{2}{9}\theta_{t-6} + \frac{3}{9} \theta_{t-7} + \frac{2}{9} \theta_{t-8} + \frac{1}{9} \theta_{t-9}. \label{eq:occ}
\end{eqnarray}
The convoluted operational cost $\theta_t^c$ cannot be observed directly, instead it is modified by the second sub-dynamic, called mis-calibration, and finally subject to observation noise.
The motivation for this dynamical behavior is that it is non-linear, it depends on more than one influence, 
and it is delayed and blurred.
All those effects have been observed in industrial applications, like the heating process observable during combustion. In Figure~\ref{fig:oc_dyn} we give an example trajectory of the convolution process over a rollout of 200 time steps. The delayed and blurred relations between operational cost $\theta_t$ and the convoluted costs $\theta_t^c$ are clearly visible.

\subsection{Dynamics of mis-calibration}
\label{sec:Dynamics-of-mis-calibration}

The sub-dynamics of mis-calibration are influenced by external driver setpoint $p$ and steering shift $h$.
The goal is to reward an agent to oscillate in $h$ in a pre-defined frequency around a specific operation point determined by setpoint $p$. Thereby, the reward topology is inspired by an example from quantum physics, namely Goldstone's "Mexican hat" potential.

In the first step, setpoint $p$ and shift $h$ are combined to an effective shift $h^\text{e}$ calculated by:
\begin{equation}
h^\text{e} = \max\left(-1.5,\min\left(1.5,\frac{h}{20} - \frac{p}{50} -1.5\right)\right).
\end{equation}
Effective shift influences three latent variables, which are domain $\delta$, response $\psi$, and direction $\phi$. 
Domain $\delta$ can enter two discrete states, which are \textit{negative} and \textit{positive}, represented by integer values $-1$ and $+1$, respectively. 
Response $\psi$ can enter two discrete states, which are \textit{disadvantageous} and \textit{advantageous}, represented by integer values $-1$, and $+1$, respectively. 
Direction $\phi\in\{-6,-5,\ldots,6\}$ is a discrete index variable, yielding the position of the current optima in the mis-calibration penalty space.

\begin{figure}
	\centering
	\includegraphics[width=.5\textwidth]{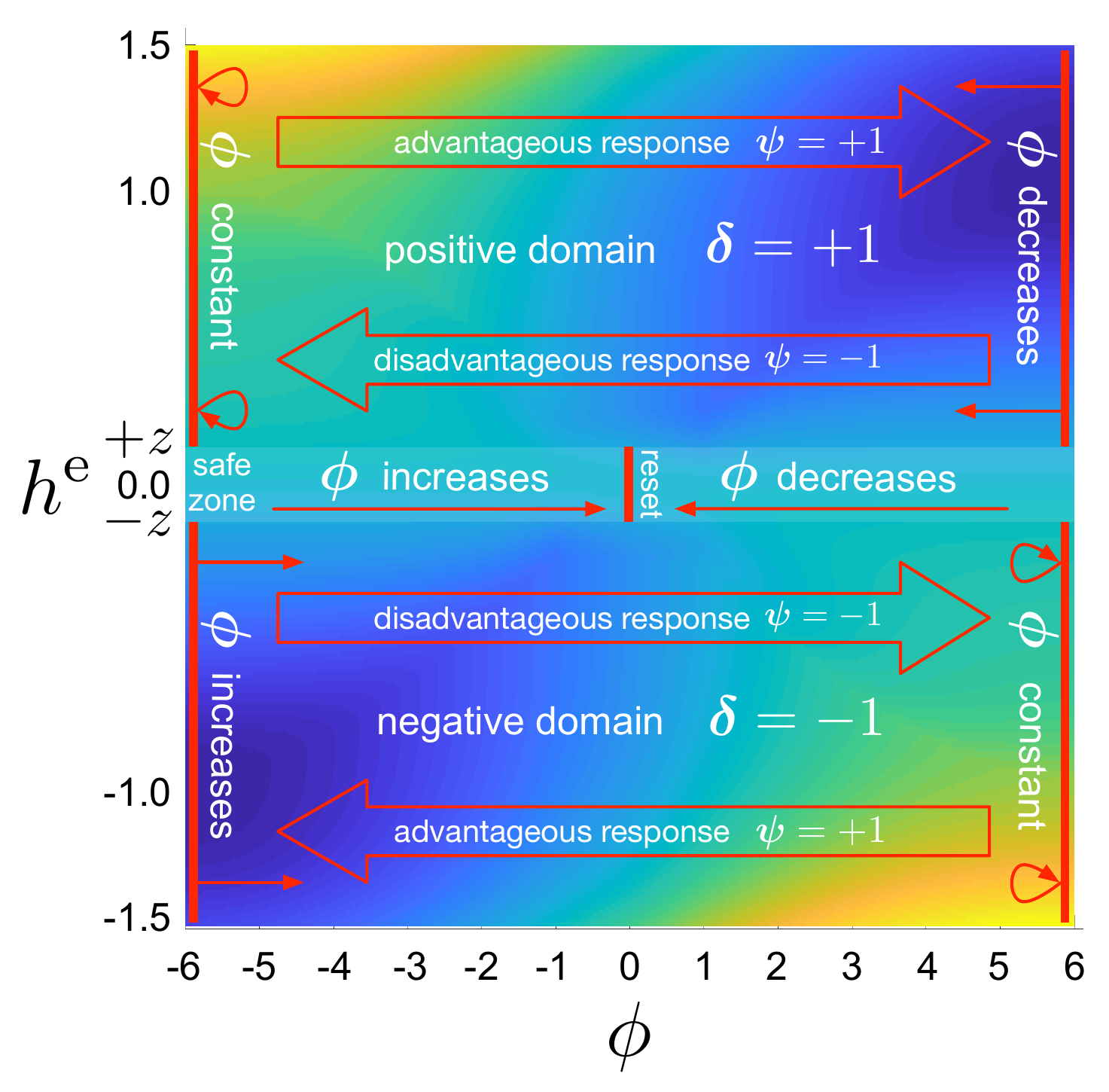}
	\caption{
		Visual description of the mis-calibration dynamics.
		Blue color represents areas of low penalty (-1.00), while yellow color represents areas of high penalty (1.23).
		If the policy keeps $h^\text{e}$ in the so-called safe zone, $\phi$ is driven towards 0 stepwise.
		When $\phi=0$ is reached the mis-calibration dynamics are reset, i.e., domain $\delta=1$ and response $\psi=1$.
		The policy is allowed to start the rotation cycle at any time by leaving the safe zone and entering the \textit{positive} or the \textit{negative} domain.
		Consider \textit{positive} domain $\delta=1$:
		After initially leaving the safe zone, response is in the state \textit{advantageous}, i.e., $\phi$ is  increased stepwise.
		The upper right area is a reversal point for $\phi$.
		As soon as $\phi=6$, response switches from \textit{advantageous} $\psi=1$ to \textit{disadvantageous} $\psi=-1$.
		In the subsequent time steps $\phi$ is decreased until either the policy brings $h^\text{e}$ back to the safe zone or $\phi$ reaches the left boundary at -6.
		If the latter occurs, phi is kept constant at -6, i.e., the policy yields a high penalty in each time step.
		Since the mis-calibration dynamics are symmetric around $(\phi,h^\text{e})=(0,0)$, opposite dynamics are applied in the \textit{negative} domain at the lower part of the plot.
	}
	\label{fig:mis_calibration}
\end{figure}

Figure~\ref{fig:mis_calibration} is a visualization of the mis-calibration dynamics introduced in equation form in the following paragraphs.
In each time step $t\rightarrow t+1$ the mis-calibration dynamics are transitioned starting with $\delta$ and $\psi$ as follows:
\begin{equation}
\hat{\delta}_{t+1} =
\begin{cases}
\delta_t & \text{if}\quad |h^\text{e}|\le z\\
\sgn(h^\text{e})  & \text{else},
\end{cases}
\end{equation}
\begin{equation}
\hat{\psi}_{t+1} =
\begin{cases}
1 & \text{if}\quad \delta_t\ne \hat{\delta}_{t+1}\\
\psi_t  & \text{else},
\end{cases}
\end{equation}
where safe zone $z$ (area in the center of Figure~\ref{fig:mis_calibration}) is calculated using $z=\sin(\pi\cdot15/180)/2\approx0.1309$.
Note that the policy itself is allowed to decide when to leave the safe zone.

In the next step, direction index $\phi$ is updated accordingly:
\begin{equation}
\hat{\phi}_{t+1}=\phi_t+\Delta\phi_{t+1}, \text{with}
\end{equation}
\begin{equation}
\Delta\phi_{t+1} =
\begin{cases}
-\sgn(\phi_t) & \text{if}\quad |h^\text{e}|\le z\\
0 & \text{if}\quad |h^\text{e}|> z \land \phi_t=-6\hat{\delta}_{t+1}\\
\hat{\psi}_{t+1}\cdot\sgn(h^\text{e})  & \text{else}.
\end{cases}
\end{equation}
The first option realizes the return of $\phi$ if the policy returns into the safe zone.
The second option stops the rotation if $\phi$ reaches the opposite domain bound (upper left and lower right area in Figure~\ref{fig:mis_calibration}).
The third option implements the cyclic movement of $\phi$ depending on response $\psi$ and the direction of effective shift $h^\text{e}$.

If, after this update, the absolute value of direction index $\phi$ reaches or exceeds the predefined maximum index of 6 (upper right and lower left area in Figure~\ref{fig:mis_calibration}), response enters state \textit{disadvantageous} and index $\phi$ is turned towards 0.
\begin{equation}
\doublehat{\psi}_{t+1} =
\begin{cases}
-1 & \text{if}\quad |\hat{\phi}_{t+1}|\ge 6\\
\hat{\psi}_{t+1}  & \text{else}.
\end{cases}
\end{equation}
\begin{equation}
\phi_{t+1} =
\begin{cases}
12-((\hat{\phi}_{t+1}+24) \mod 24) & \text{if}\quad |\hat{\phi}_{t+1}|\ge6\\
\hat{\phi}_{t+1}  & \text{else}.
\end{cases}
\end{equation}

In the final step of the mis-calibration state transition, it is checked if effective shift $h^\text{e}$ has returned to safe zone $z$ while at the same time direction index $\phi$ has completed a full cycle (reset area in the center of Figure~\ref{fig:mis_calibration}). 
If this is the case, domain $\delta$ and response $\psi$ are reset to their initial states \textit{positive} and \textit{advantageous}, respectively:
\begin{equation}
\delta_{t+1} =
\begin{cases}
1 & \text{if}\quad \phi_{t+1}=0 \land |h^\text{e}|\le z\\
\hat{\delta}_{t+1}  & \text{else}.
\end{cases}
\end{equation}
\begin{equation}
\psi_{t+1} =
\begin{cases}
1 & \text{if}\quad \phi_{t+1}=0 \land |h^\text{e}|\le z\\
\doublehat{\psi}_{t+1}  & \text{else}.
\end{cases}
\end{equation}
Note that in this state the policy can again decide to start a new cycle (positive or negative direction) or to remain in state $\phi=0$.

\begin{figure*}[th!]
	\centering
	\subfloat[Optimal policy]{
		\includegraphics[width=.3\textwidth]{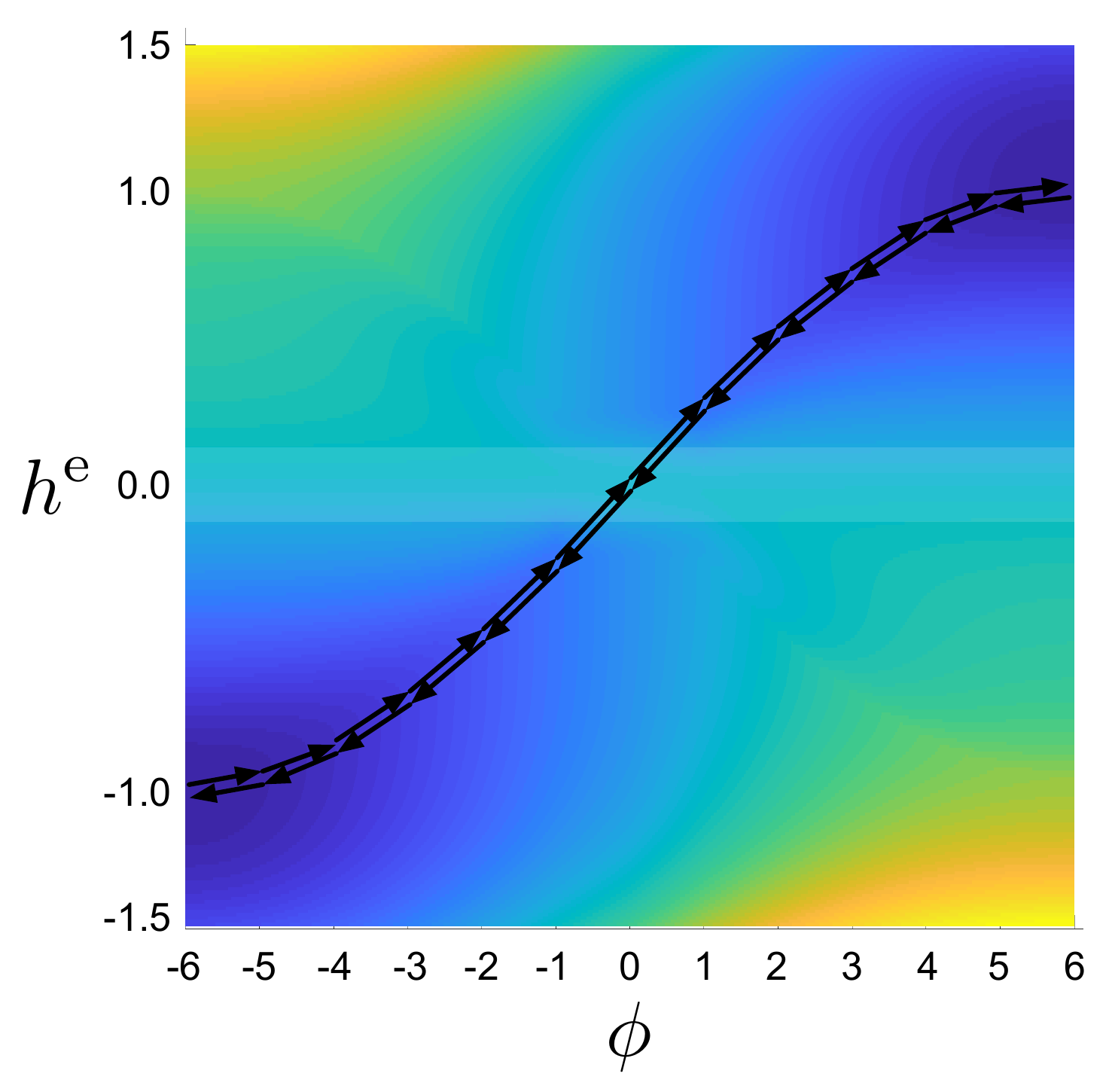}
		\label{fig:shift_opt}
	}
	\subfloat[Suboptimal policy]{
		\includegraphics[width=.3\textwidth]{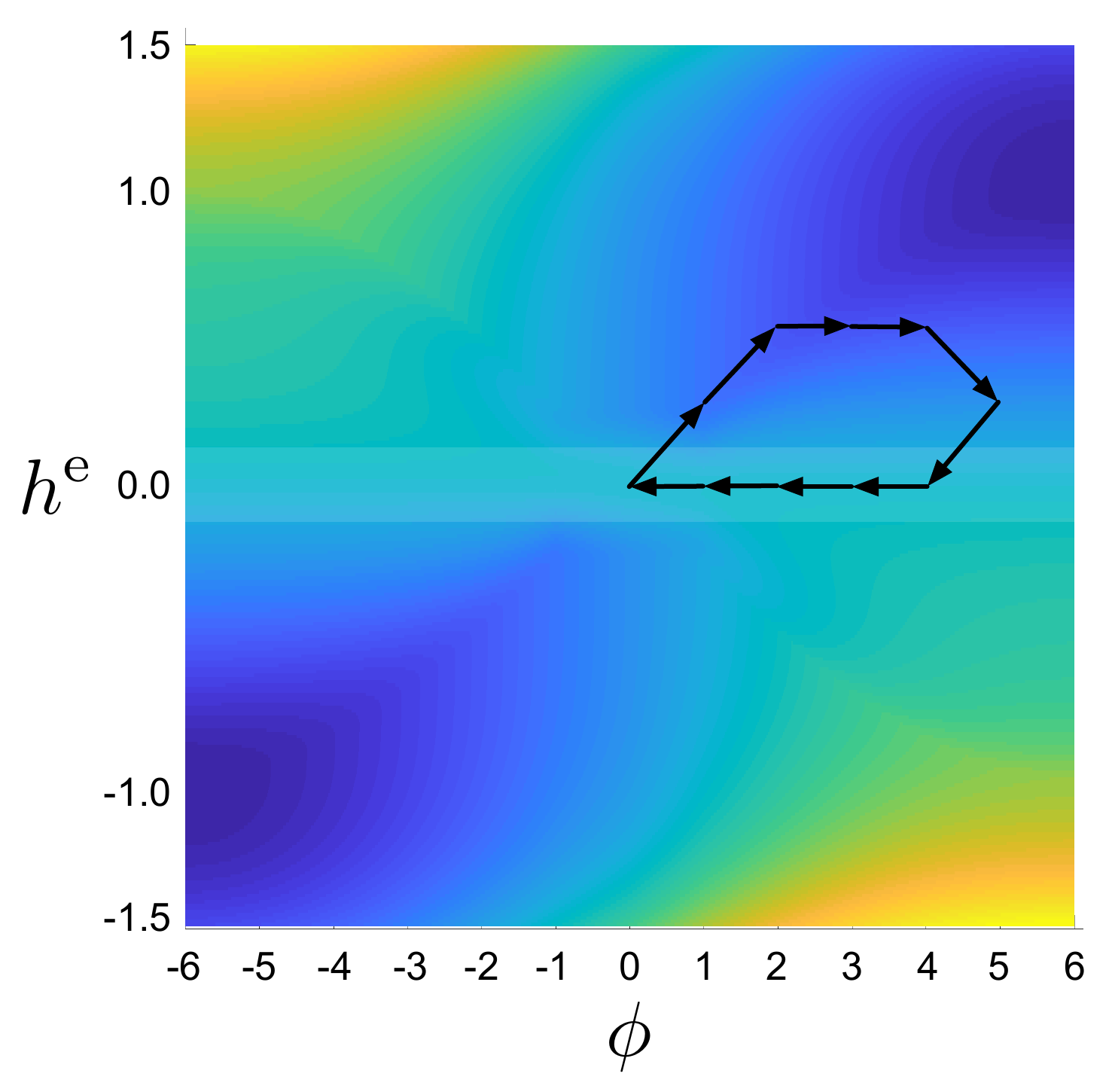}
		\label{fig:shift_sub}
	}
	\subfloat[Bad policy]{
		\includegraphics[width=.3\textwidth]{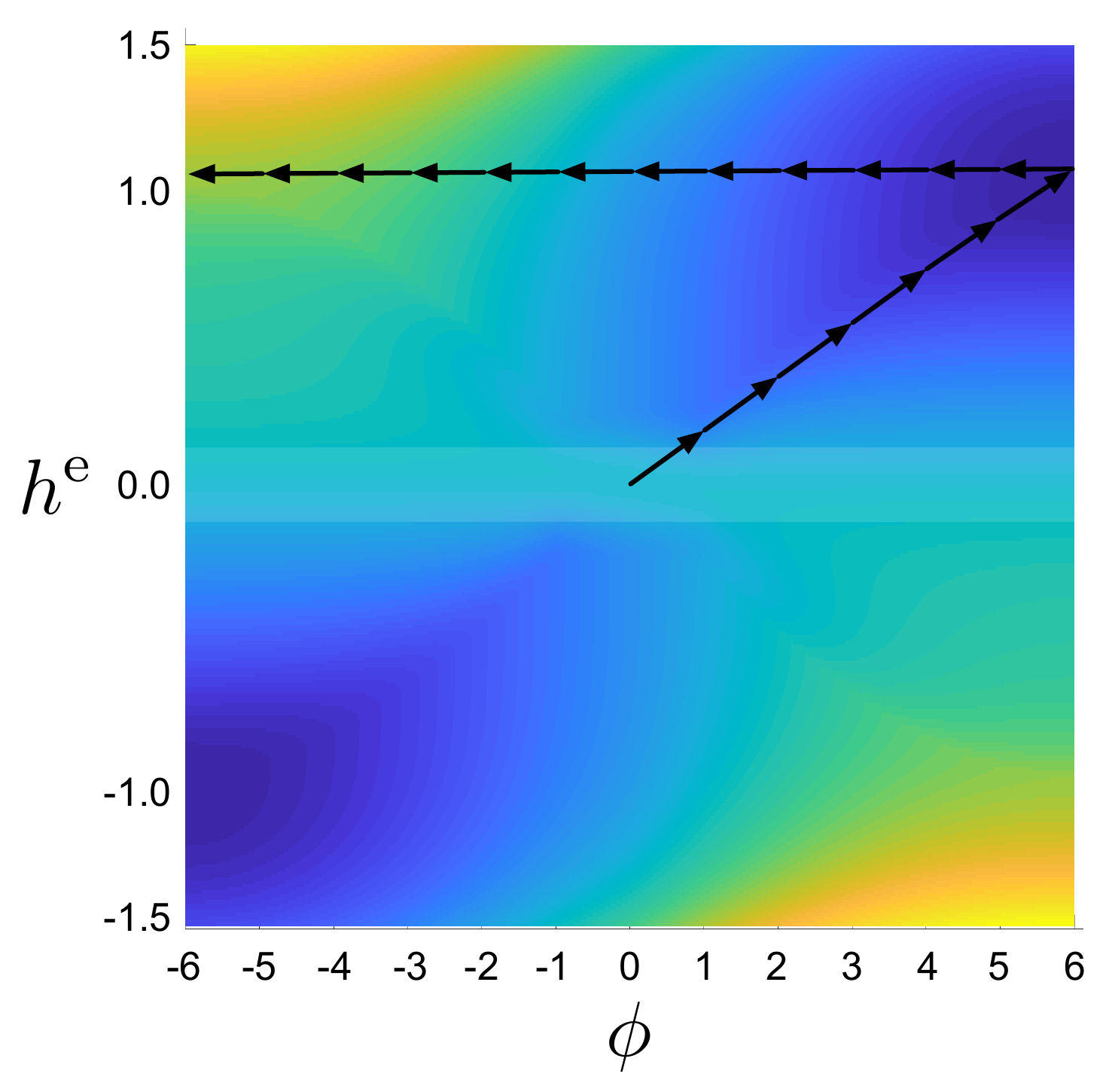}
		\label{fig:shift_bad}
	}
	\caption{
		Comparison of three mis-calibration policies.
		Depicted is a visual representation of the Goldstone potential based function $m(\phi,h^\text{e})$.
		Areas yielding high penalty are colored yellow, while areas yielding low penalty are colored blue.
		The highlighted area in the center depicts the safe zone from $-z$ to $z$.
		\protect\subref{fig:shift_opt}~A policy which maintains $h^\text{e}$ such that a sine-shaped trajectory is generated yields lowest penalty.
		Note that the policy itself starts the rotation cycle at any time by leaving the safe zone.
		After returning to the safe zone, while at the same time $\phi=0$, the dynamics are reset and a new cycle can be initiated at any following time step in positive or negative direction.	
		\protect\subref{fig:shift_sub}~The depicted policy starts initiating the rotation cycle by leaving the safe zone, but returns after six steps.
		After this return $\phi$ is decreased in four steps back to 0.
		Subsequently, the dynamics are reset.
		This policy yields lower penalty compared to a constant policy that remains in the safe zone the whole time.
		\protect\subref{fig:shift_bad}~The depicted policy approaches one of the global optima of $m(\phi,h^\text{e})$ by directly leaving the safe zone $z$ by constantly increasing $h^\text{e}$.
		Subsequently, it remains at this point.
		However, the rotation dynamic yields a steady decrease in $\phi$ after reaching the right boundary at $\phi=6$.
		This decrease "pushes" the agent to the left, i.e., the penalties received are increased from step to step.
		After reaching the left boundary at $\phi=-6$, the dynamics remain in this area of high penalty.
		Note that the policy could bring the dynamics back to the initial state $\phi=0$ by returning to $h^\text{e}<z$.
		This benchmark property ensures that the best constant policies are the ones which remain in the safe zone.
	}
	\label{fig:policies}
\end{figure*}

The penalty landscape of mis-calibration is computed as follows.
Based on the current value of $\phi$, the penalty function $m(\phi,h^\text{e})$ computes the performance of maintaining shift in the beneficial area.
The penalty function $m$ is defined as a linearly biased Goldstone potential computed by
\begin{equation}
m = -\alpha\omega^2+\beta\omega^4+\kappa\rho^\text{s}\omega.
\end{equation}
The definition of radius $\omega$ can be found in Appendix~\ref{appendix:goldstone}.
From direction index $\phi$ the sine of direction angle $\rho$ is calculated as follows:
\begin{equation}
\rho^\text{s} = \sin\left(\frac{\pi}{12}\phi\right).
\end{equation}
Note that this sine function represents the optimal policy for the mis-calibration dynamics.
Exemplary policy trajectories through the penalty landscape of mis-calibration are depicted and described in Figure~\ref{fig:policies}.

The resulting mis-calibration $m_t$ is added to the convoluted operational cost $\theta^\text{c}_t$, giving $\hat{c}_t$,
\begin{equation}
\hat{c}_t = \theta^\text{c}_t + 25 m_t.
\end{equation}
Before being observable as consumption $c_t$, the modified operational cost $\hat{c}_t$ is subject to heteroscedastic observation noise
\begin{equation}
c_t = \hat{c}_t + \rm{gauss}(0,1+0.02\,\hat{c}_t)\,,
\end{equation}
i.e., a Gaussian noise with zero mean and a standard deviation of $\sigma=1+0.02\,\hat{c}_t$. 
In Figure~\ref{fig:cons_dyn} we show in an example rollout of 200 steps how both convoluted operational cost $\theta^c_t$ and mis-calibration $m_t$ affect consumption $c_t$.

\subsection{Dynamics of fatigue}
The sub-dynamics of fatigue are influenced by the same variables as the sub-dynamics of operational cost, i.e., setpoint $p$, velocity $v$, and gain $g$. 
The IB is designed in such a way that, when changing the steerings velocity $v$ and gain $g$ as to reduce the operational cost, fatigue will be increased, leading to the desired multi-criterial task, with two reward-components showing opposite dependencies on the actions.
The basic fatigue $f^\text{b}$ is computed as 
\begin{equation}
f^\text{b} = \max\left(0,\frac{30000}{5\,v + 100} - 0.01\, g^2\right).
\end{equation}
From basic fatigue $f^\text{b}$, fatigue $f$ is calculated by
\begin{equation}
f = f^\text{b} (1 + 2 \alpha)/3\,,
\end{equation}
where $\alpha$ is an amplification.
The amplification depends on two latent variables $\mu^\text{v}$ and $\mu^\text{g}$,
effective velocity $v^\text{e}$, and effective gain $g^\text{e}$. 
Furthermore, it is affected by noise,
\begin{equation}\label{eq:alpha}
  \alpha =
  \begin{cases}
    \frac{1}{1+\exp(-{\rm gauss(2.4,0.4)})} & \text{if}\quad \max(\mu^\text{v},\mu^\text{g})=1.2\\
    \max(\eta^\text{v},\eta^\text{g})       & \text{else}.
  \end{cases}
\end{equation}
{In Eq.~\eqref{eq:alpha} we see that $\alpha$ can undergo a bifurcation if one of the latent variables $\mu^\text{v}$ or $\mu^\text{g}$ reaches  a value of $1.2$. 
In that case, $\alpha$ will increase and lead to higher fatigue, affecting the reward negatively.

The noise components $\eta^\text{v}$ and $\eta^\text{g}$, as well as the latent variables $\mu^\text{v}$ and $\mu^\text{g}$, depend on effective velocity $v^\text{e}$, and effective gain $g^\text{e}$.
These are calculated by setpoint-dependent transformation functions
\begin{eqnarray}
{\rm T}^\text{v}(v,g,p) & = & \frac{g + p + 2}{v - p + 101},\\
{\rm T}^\text{g}(g,p)   & = & \frac{1}{g + p + 1}\,.
\end{eqnarray}

Based on these transformation functions, effective velocity $v^\text{e}$ and effective gain $g^\text{e}$ are computed as follows:
\begin{eqnarray}
v^\text{e} & = & \frac{ {\rm T}^\text{v}(v,g,p) - {\rm T}^\text{v}(0,100,p)}
               { {\rm T}^\text{v}(100,0,p) - {\rm T}^\text{v}(0,100,p)}\\
g^\text{e} & = & \frac{ {\rm T}^\text{g}(g,p) - {\rm T}^\text{g}(100,p)}
               { {\rm T}^\text{g}(0,p) - {\rm T}^\text{g}(100,p)}.
\end{eqnarray}
To compute the noise components $\eta^\text{v}$ and $\eta^\text{g}$, six random numbers are drawn from different random distributions:
$\eta^\text{ve}$ and $\eta^\text{ge}$ are obtained by first sampling from an exponential distribution with mean 0.05 and applying the logistic function to these samples afterwards, $\eta^\text{vb}$ and $\eta^\text{gb}$ are drawn from binomial distributions ${\rm Binom}(1, v^\text{e})$ and ${\rm Binom}(1, g^\text{e})$, respectively,
$\eta^\text{vu}$ and $\eta^\text{gu}$ are drawn from a uniform distribution in $[0,1]$.
Noise components $\eta^\text{v}$ and $\eta^\text{g}$ are computed as follows:
\begin{eqnarray}\label{eq:noise_ft}
\eta^\text{v} & = & \eta^\text{ve} + (1 - \eta^\text{ve}) \eta^\text{vu} \eta^\text{vb} v^\text{e}\\
\eta^\text{g} & = & \eta^\text{ge} + (1 - \eta^\text{ge}) \eta^\text{gu} \eta^\text{gb} g^\text{e}\,.
\end{eqnarray}

The latent variables $\mu^\text{v}$ and $\mu^\text{g}$ are calculated as
\resizebox{\linewidth}{!}{
  \begin{minipage}{\linewidth}
\begin{align}
  \mu^\text{v}_t & = &
  \begin{cases}
    v^\text{e} & \text{if}\quad v^\text{e} \le 0.05\\
    \min(5,1.1 \mu^\text{v}_{t-1}) & \text{if}\quad v^\text{e} > 0.05 \land \mu^\text{v}_{t-1} \ge 1.2\\
    0.9 \mu^\text{v}_{t-1} + \frac{\eta^\text{v}}{3} & \text{else},\\
  \end{cases} \label{Eq:hv} \\
  \mu^\text{g}_t & = &
  \begin{cases}
    g^\text{e} & \text{if}\quad g^\text{e} \le 0.05\\
    \min(5,1.1 \mu^\text{g}_{t-1}) & \text{if}\quad g^\text{e} > 0.05 \land \mu^\text{g}_{t-1} \ge 1.2\\
    0.9 \mu^\text{g}_{t-1} + \frac{\eta^\text{g}}{3} & \text{else}.\\
  \end{cases} \label{Eq:hg}
\end{align}
\end{minipage}}
 
The sub-dynamic of fatigue results in a value for fatigue $f$, which is relevant for the reward function (Eq.~ \ref{eq:reward}).
\begin{figure*}[t]
	\centering
	\subfloat[Fatigue dynamics]{
		\includegraphics[width=.48\textwidth]{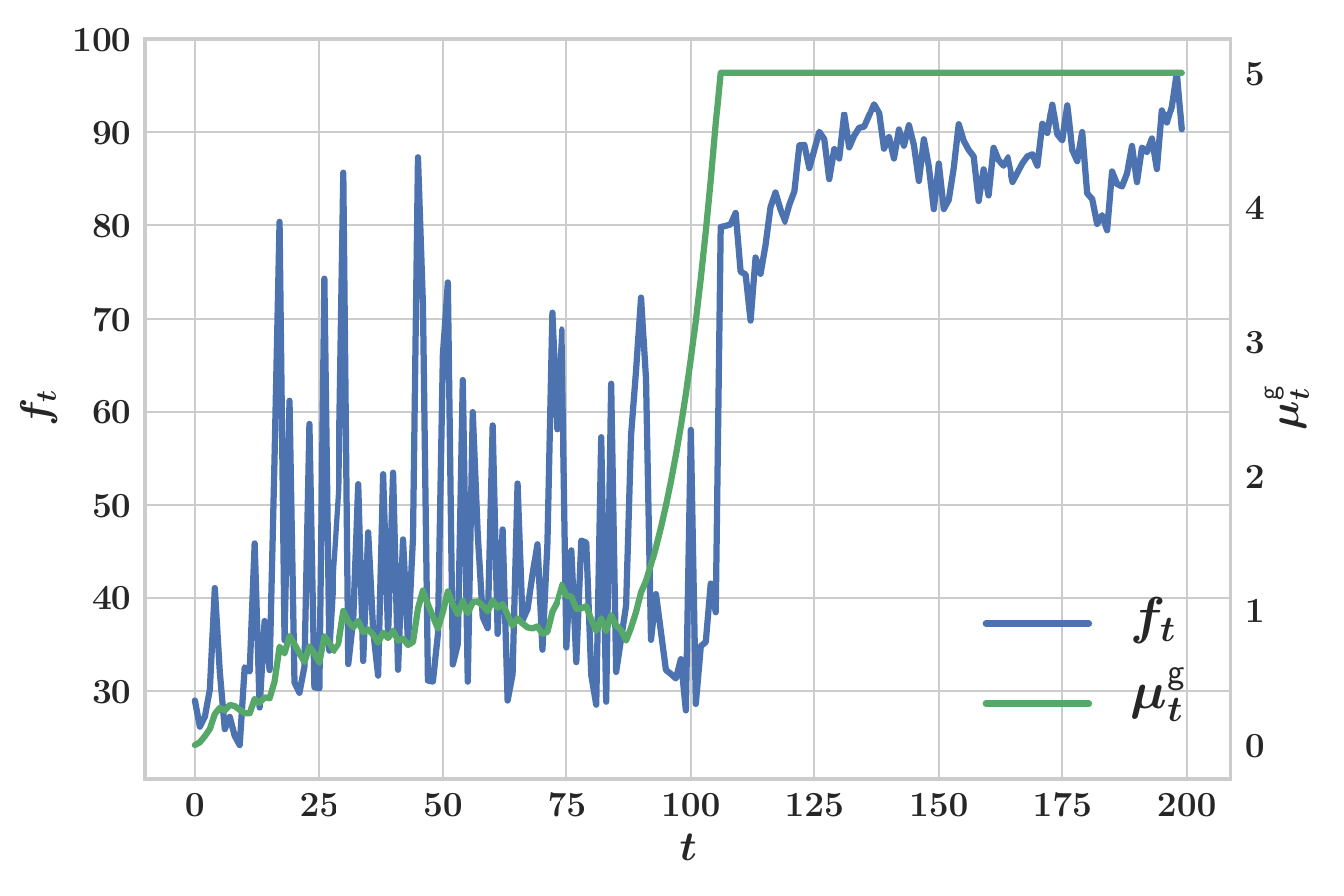}
		\label{fig:fatigue_dyn}
	}
	\subfloat[Operational cost]{
		\includegraphics[width=.48\textwidth]{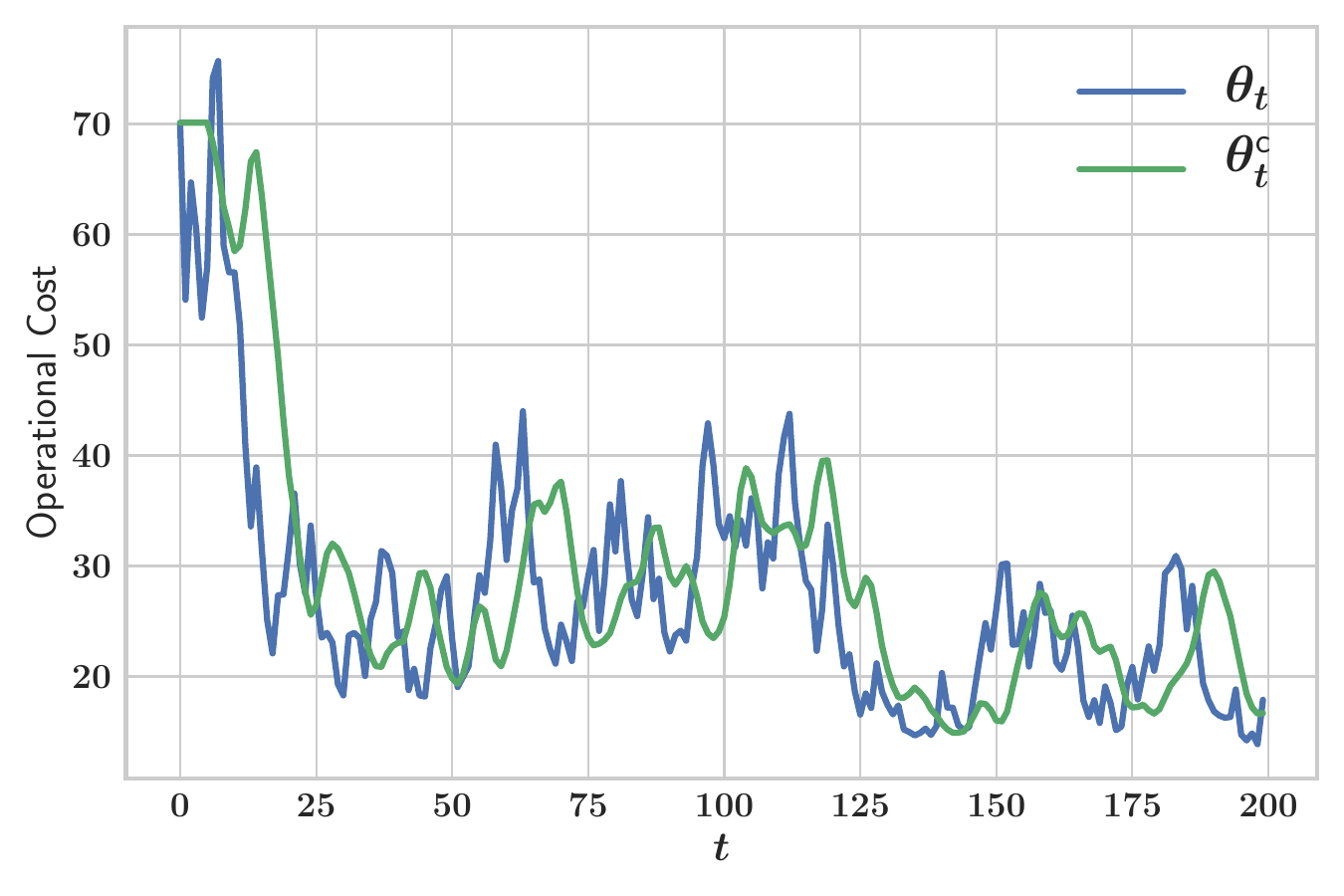}
		\label{fig:oc_dyn}
	} \\
	\subfloat[Consumption dynamics]{
		\includegraphics[width=.48\textwidth]{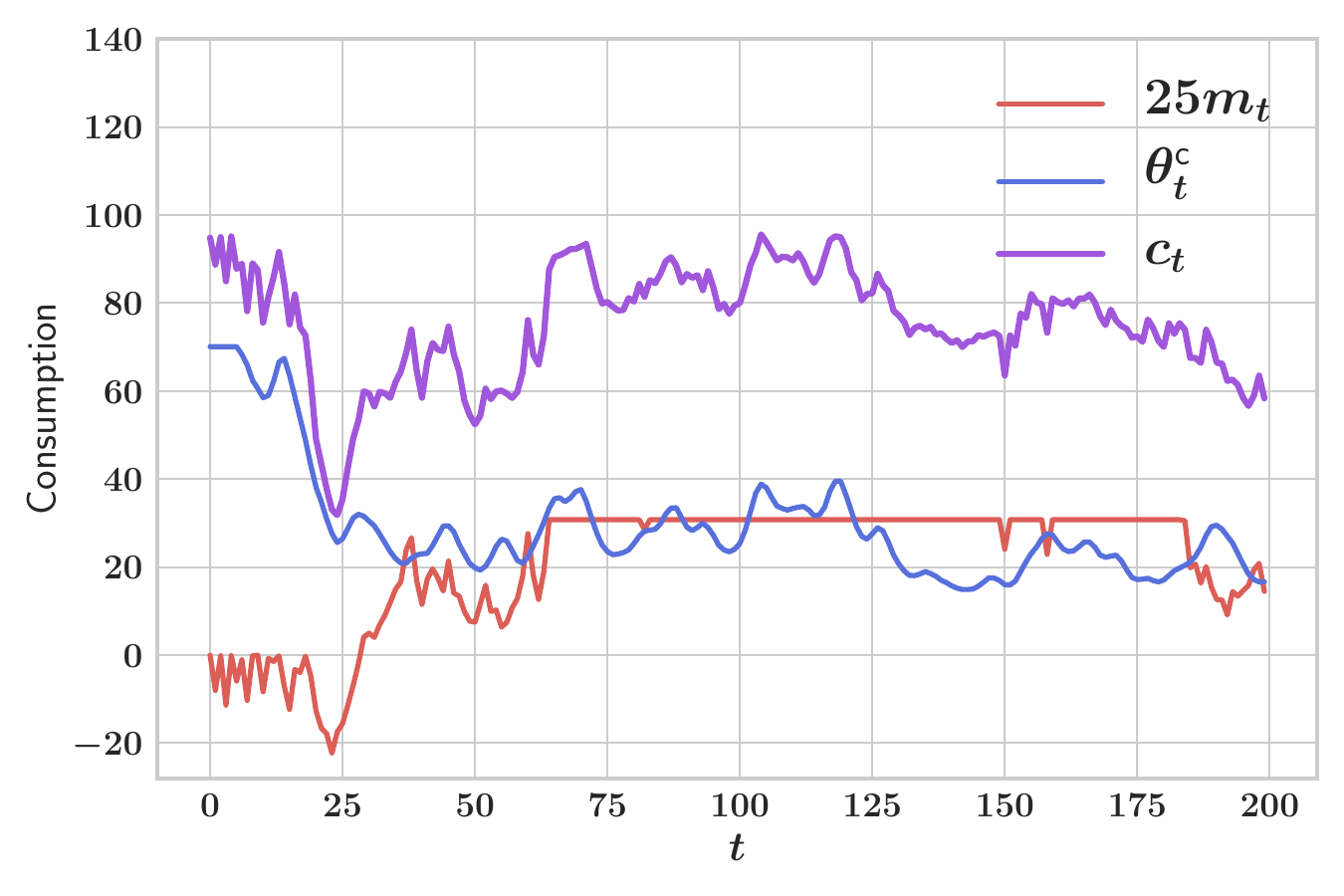}
		\label{fig:cons_dyn}
		
	}
	\subfloat[Reward dynamics]{
		\includegraphics[width=.48\textwidth]{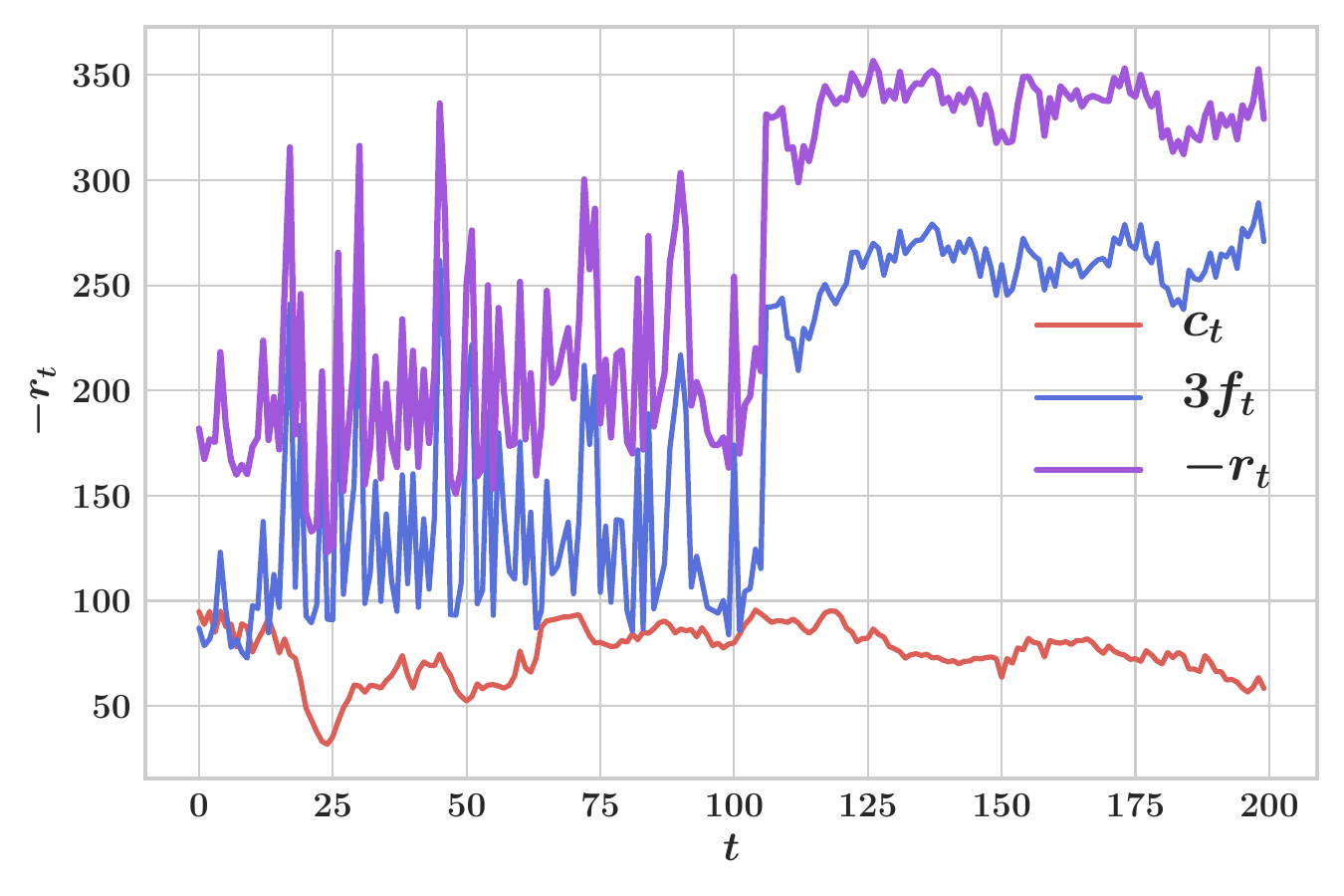}
		\label{fig:rew_dyn}
		
	}
	\caption{
		Visualization of relevant variables of the IB in a rollout using random actions over 200 time steps.
		\protect\subref{fig:fatigue_dyn}: Shown are latent variable $\mu^\text{g}$ and fatigue $f$. 
		As seen in Eq.~\eqref{eq:alpha}, the latent variable can lead to a bifurcation of the dynamics. 
		In the scenario shown at $t=90$, we observe the beginning of a runaway effect that originates from the second case in Eq.~\eqref{Eq:hg}.
		\protect\subref{fig:oc_dyn}: Shown are operational cost $\theta_t$ and convoluted sigma $\theta_t^\text{c}$ given by Eq.~\eqref{eq:occ}. 
		At around $t=10$ the delayed effect of the convolution is clearly visible: $\theta_t$ decreases sharply while $\theta_t^\text{c}$ is still ascending. 
		\protect\subref{fig:cons_dyn}: Shown is the composition of visible consumption $c(t)$ (purple) by the two components $\sigma^c_t$ and mis-calibration $m_t$. 
		\protect\subref{fig:rew_dyn}: Shown is the composition of final negative reward $-r_t$ by its two components, fatigue (blue) and consumption (red). 
		In this case, the runaway effect from Figure~\protect\subref{fig:fatigue_dyn} has the most prominent effect on the reward signal.
	}
	\label{fig:dynamics}
\end{figure*}
An example interplay of the components of the fatigue dynamics is visualized in Figure~\ref{fig:fatigue_dyn}. From $t=0$ up to $t=80$ we see the effect of the noise components described in Eq.~\eqref{eq:noise_ft}: the combination of binomial and exponential noise components yields heterogeneous spike-like behavior. 
From $t=80$ to $t=100$ we observe a self-amplifying process in $\mu^\text{g}$. 
This self-amplification originates from the second case of Eq.~\eqref{Eq:hg}. 
At around $t=100$, the fatigue dynamics rapidly change towards higher, less noisy regions. 
This change originates from the bifurcation in $\alpha$ in Eq.~\eqref{eq:alpha}, which we pointed out earlier.

\subsection{Setpoint dynamics}
\label{setpoint_dyn}

Setpoint $p$ can either be kept constant or it is variable over time. 
In the variable setting, it will change by a constant value $b$ over a fixed period of $l$ time steps in the benchmark dynamics. 
Afterwards, a new sequence length and change rate is determined.

We sample sequence length $l$ uniformly from $\mathcal{U}\{1,100\}$ and draw rate $b$ from a mixture of a uniform distribution $\mathcal{U}(0,1)$ and a delta distribution $\delta(x)$ with weighting probabilities 0.9 and 0.1. For each time step $t+1$ we update the setpoint according to:
\begin{align}
p_{t+1} &=\max(0,\min(100,p_t + b_{t+1})),\\
b_{t+1}  &= \begin{cases} 
			-b_t &\text{if } (p_t=0 \lor p_t=100) \land z < 0.5\\
			b_t &\text{else},
        \end{cases}
\end{align}
where $z\sim\mathcal{U}(0,1)$ will flip change rate $b$ with a probability of 50\% if the setpoint reaches one of the two bounds at $p=0$ or $p=100$. 
Note that the equation above produces piecewise linear functions of constant change. 
We visualized four example trajectories in Figure~\ref{fig:setpoints}.

\begin{figure}[t]
\centering
\includegraphics[width=\linewidth]{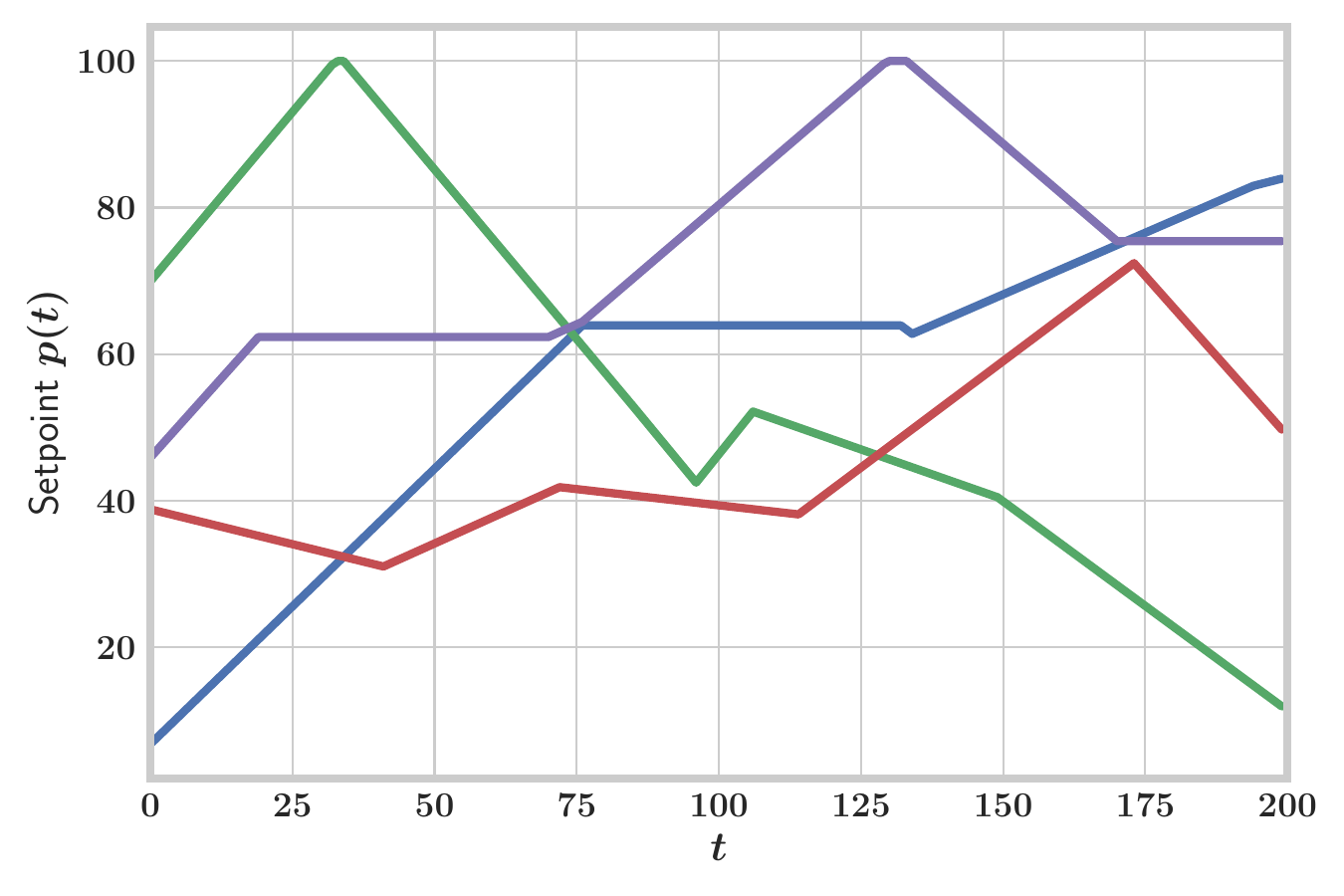}
\caption{Four example trajectories of the setpoint of the IB in a variable setpoint setting.}
\label{fig:setpoints}
\end{figure}

%% file: setup.tex
\section{Experimental Prototypes}
\label{sec:Experimental_setup}

The IB aims at being realistic in the sense, that it includes a variety of aspects that we found to be vital in industrial applications. 
In this section we want to outline prototypes of experimental settings that include key aspects present in industrial applications.

\subsection{Batch Reinforcement Learning}
In this setting, we are given an initial batch of data $D$ from an already-running system and are asked to find a better (ideally near-optimal) policy. 

The learners task is therefore to return a policy $\pi(\mathbf{s}_t)$ that can be deployed on the system at hand, solely
based on the information provided by the batch~\cite{lange2012batch}. 
These scenarios are common in real-world industry settings where exploration is usually restricted to avoid possible damage to the system. 

Two scenarios using the IB for batch RL experiments are described subsequently. 

\subsubsection*{Random exploration}
In this setting, we generate a batch of state transitions using a random behavior policy, for instance by sampling action proposals from a uniform distribution. 
Example instances of these settings can be found in~\cite{depeweg2016learning} and \cite{hein2017batch}.

In the cited examples, the benchmark is initialized for ten different setpoints $p\in\{10,20,\ldots,100\}$ with the latent variables in their default values and the three steering variables at 50 each. 
Then, for each setpoint $p$ the behavior policy is applied on the benchmark for 1,000 time steps, resulting in a total of 10,000 recorded state transitions. 
This process can be repeated to study the performance using different batch sizes.

For evaluation, the system is either initialized to its start settings~\cite{hein2017batch}, or at a random place in state space~\cite{depeweg2016learning}, at which point the policy drives the system autonomously.

\subsubsection*{Safe behavior policy}
In real industrial settings, we seldom will run a fully random policy on the system at hand. 
A more realistic setting is that we have a batch of data generated by a safe, but suboptimal, behavior policy $\pi_b$ with limited randomness.
In this setting, the task is to improve $\pi_b$. 
Unlike in the random exploration setting, the difficulty here is that large parts of the state space will be unavailable in the batch. 
The batch of data will likely contain more information of specific areas in state space and few information everywhere else.
An example experiment can be found in~\cite{depeweg:17}.

\subsection{Transfer Learning}
A common situation in industrial control is that we have data from different industrial systems, or data from one industrial system that operates in different contexts. 
We expect that each instance will behave similarly on a global level, while we can expect significant deviations on a low level.

In the IB, this is realized by the setpoint $p$, a hyperparameter of the dynamics.
Each value of $p \in [0,\ldots,100]$ will define a different stochastic system, where the dissimilarity of two systems grows with the distance in $p$.

In transfer learning, we want to transfer our knowledge from system A to system B. 
For example, suppose we have a large batch $D_1$ of state transitions from the IB with $p=50$. 
We also have a small batch of state transition $D_2$ with $p=75$. 
If our goal is to learn a good model for a system with $p=75$, the challenge of transfer learning is how to efficiently incorporate the batch $D_1$ to improve learning. 
An example instance of this setup, albeit using pendulum dynamics, can be found in~\cite{spieckermann2015exploiting}.

%% file: conclusion.tex
\section{Conclusion}

This paper introduced the IB, a novel benchmark for RL, inspired by industrial control. 
We have shown that it provides a useful addition to the set of existing RL benchmarks due to its unique combination of properties. 
Furthermore, we outlined prototype experimental setups relevant for industrial control. 
Our contributions are a step towards enabling other researchers to study RL in realistic industrial settings to expand the economical and societal impact of machine learning.

%% file: states.tex
\section{State Description}
\label{appendix:states}

Only a part of the state variables is observable. 
This observation vector is also called observable state, but one has to keep in mind, that it does not fulfill the Markov property.
The observation vector $o_t$ at time $t$ comprises current values of velocity $v_t$, gain $g_t$, shift $h_t$, setpoint $p_t$, consumption $c_t$, and fatigue $f_t$.

The preferred minimal Markovian state fulfills the Markov property with the minimum number of variables.
It comprises 20 values. 
These are the observation vector (velocity $v_t$, gain $g_t$, shift $h_t$, setpoint $p_t$, consumption $c_t$, and fatigue $f_t$) plus some latent variables of the sub-dynamics.
The sub-dynamics of operational cost add a list of previous operational costs, $\theta_{t-i}$ with $i\in {1,\cdots,9}$.
Note that the current operational cost $\theta_t$ is not part of this state definition.
It would be redundant, as it can be calculated by $v_t$, gain $g_t$, and setpoint $p_t$.
The sub-dynamics of mis-calibration need three additional latent variables, $\delta$, $\psi$, and $\phi$, (Section~\ref{sec:Dynamics-of-mis-calibration}).
The sub-dynamics of fatigue add 2 additional latent variables $h^\text{v}$ and $h^\text{g}$,
(Eq.~\eqref{Eq:hv} and \eqref{Eq:hg}).

\begin{table}[!ht]
	\begin{center}
		\begin{tabular}{cclll}
			\toprule
			& & text name or description & symbol\\
			\midrule
			\multirow{4}{*}{\begin{turn}{90}------------------------ Markovian state ------------------------\end{turn}}& \multirow{4}{*}{\begin{turn}{90}-- Observables --\end{turn}} & setpoint   & $p_t$\\
			& & velocity    & $v_t$\\
			& & gain        & $g_t$\\
			& & shift       & $h_t$\\
			& & consumption & $c_t$\\
			& & fatigue     & $f_t$\\
			\cmidrule{2-5}
			& & operational cost at $t-1$ & $\theta_{t-1}$\\
			& & operational cost at $t-2$ & $\theta_{t-2}$\\
			& & operational cost at $t-3$ & $\theta_{t-3}$\\
			& & operational cost at $t-4$ & $\theta_{t-4}$\\
			& & operational cost at $t-5$ & $\theta_{t-5}$\\
			& & operational cost at $t-6$ & $\theta_{t-6}$\\
			& & operational cost at $t-7$ & $\theta_{t-7}$\\
			& & operational cost at $t-8$ & $\theta_{t-8}$\\
			& & operational cost at $t-9$ & $\theta_{t-9}$\\
			& & $1^{\rm st}$ latent variable of mis-calibration & $\delta$\\
			& & $2^{\rm nd}$ latent variable of mis-calibration & $\psi$ &\\
			& & $3^{\rm rd}$ latent variable of mis-calibration & $\phi$ &\\
			& & $1^{\rm st}$ latent variable fatigue & $\mu^\text{v}$ &\\
			& & $2^{\rm nd}$ latent variable fatigue & $\mu^\text{g}$ &\\
			\bottomrule	
		\end{tabular}
	\end{center}
	\caption{IB Markovian state.}
\end{table}

%% file: goldstone.tex
\section{Goldstone Potential Based Equations}
\label{appendix:goldstone}

The resulting penalty of the mis-calibration reward component is computed by adopting a so-called linearly biased Goldstone potential.
The following constants are pre-defined to subsequently compute the respective penalty:
\begin{align}
	\epsilon &= \frac{\sqrt[3]{1+\sqrt{2}}}{\sqrt{3}}\approx0.7745,\\
	\zeta &= \epsilon+\frac{1}{3\epsilon}\approx1.2048,\\
	\lambda &= 2\zeta^2-\zeta^4+8\sqrt{\frac{2}{27}}\zeta\approx3.4193,\\
	\alpha &= \frac{2}{\lambda}\approx0.5849,\\
	\beta &= \frac{1}{\lambda}\approx0.2924,\\
	\kappa &= \frac{-8\sqrt{\frac{2}{27}}}{\lambda}\approx-0.6367.
\end{align}

Given effective shift $h^\text{e}$ and the sine of direction angle $\rho$, which is denoted as $\rho^\text{s}$, function $\omega(\rho^\text{s},h^\text{e})$ is computed using the following set of equations:
\begin{align}
\omega &=
\begin{cases}
(\frac{h^\text{e}|r^\text{min}|)}{|r^\text{opt}|} & \text{if}\quad |h^\text{e}|\le|r^\text{opt}|\\
\sgn(h^\text{e})\cdot \hat{\omega} & \text{else},
\end{cases}
\\
\hat{\omega} &= |r^\text{min}|+\frac{2-|r^\text{min}|}{(2-|r^\text{opt}|)^{\doublehat{\omega}}}\cdot(|h^\text{e}|-|r^\text{opt}|)^{\doublehat{\omega}},
\\
\doublehat{\omega} &= \frac{2-|r^\text{opt}|}{2-|r^\text{min}|},
\\
r^\text{opt} &= \varrho\cdot\max(|\rho^\text{s}|,2z),
\\
r^\text{min} &= 
\begin{cases}
u+\frac{1}{3u} & \text{if}\quad q<-\sqrt{1/27}\\
\varrho\sqrt{\frac{4}{3}}\cos\left(\frac{1}{3}\cdot\text{acos}(-q\sqrt{27})\right)  & \text{else},
\end{cases}
\\
u &= \sqrt[3]{-\varrho q+\sqrt{q^2-\frac{1}{27}}},
\\
q &= \frac{\kappa|\rho^\text{s}|}{8\beta},
\\
\varrho &= \sgn(\rho^\text{s}).
\end{align}